%% file: ged.tex
\documentclass[10pt]{scrartcl}

\input{config}

\begin{document}

\title{Geometry of Graph Edit Distance Spaces}
\author{Brijnesh J.~Jain \\
       Technische Universit\"at Berlin, Germany\\
       e-mail: brijnesh.jain@gmail.com}
            
\date{}
\maketitle

\begin{abstract} 
In this paper we study the geometry of graph spaces endowed with a special class of graph edit distances. The focus is on geometrical results useful for statistical pattern recognition. The main result is the Graph Representation Theorem. It states that a graph is a point in some geometrical space, called orbit space. Orbit spaces are well investigated and easier to explore than the original graph space. We derive a number of geometrical results from the orbit space representation, translate them to the graph space, and indicate their significance and usefulness in statistical pattern recognition.
\end{abstract}

\section{Introduction}

The graph edit distance is a common and widely applied distance function for pattern recognition tasks in diverse application areas such as computer vision, chemo- and bioinformatics \cite{Gao2010}. One persistent problem of graph spaces endowed with the graph edit distance is the gap between structural and statistical methods in pattern recognition \cite{Bunke2001,Duin2012}. This gap refers to a shortcoming of powerful mathematical methods that combine the advantages of structural representations under edit transformations with the advantages of statistical methods defined on Euclidean spaces. 

One reason for this gap is an insufficient understanding of the geometry of graph spaces endowed with the graph edit distance. Few exceptions towards a better understanding of graph spaces are, for example, theoretical results presented in  \cite{Bunke1997,Hurshman2015,Jain2009}. However, a sound theory towards statistical graph analysis in the spirit of \cite{Marron2014,Wang2007} for complex objects, \cite{Feragen2013,Wang2007} for tree-structured data, and \cite{Dryden1998,Kendall1984} for shapes is still missing. 

Here, we study the geometry of graph spaces with the goal to establish a mathematical foundation for statistical analysis on graphs. The basic ideas of this contribution build upon \cite{Jain2009} and are inspired by \cite{Feragen2011a,Feragen2011b,Feragen2013}. The graphs we study comprise directed as well as undirected graphs. Nodes and edges may have attributes from arbitrary sets such as, for example, real values, feature vectors, discrete symbols, strings, and mixtures thereof. We assume that graphs have bounded number of nodes.

The key result is the Graph Representation Theorem \ref{theorem:grt} formulated for graph edit kernel spaces. A graph edit kernel space is a graph space endowed with a geometric version of the graph edit distance. Theorem \ref{theorem:grt} is useful, because it provides deep insight into the geometry of graph spaces and simplifies derivation of many interesting results relevant for narrowing the gap between structural and statistical pattern recognition, which would otherwise be a complicated endeavour when done in the original graph space. 

The Graph Representation Theorem \ref{theorem:grt} states that a graph is a point in a geometrical space, called orbit space. The geometry and topology of orbit spaces is well investigated and much easier to explore than those of graph spaces. Based on the Graph Representation Theorem \ref{theorem:grt} we show that the graph space is a geodesic space, prove a weak form of the  Cauchy-Schwarz inequality and derive basic geometrical concepts such as the length, angle, and orthogonality. Then we present geometrical results from the point of view of a generic graph. One result is the weak version of Theorem \ref{theorem:grt}. It  states that the graph space looks like a convex polyhedral cone from the perspective of a generic graph. This result is useful, because it supports geometric intuition for deriving further results. Finally, we indicate the significance and usefulness of the derived geometrical results for statistical pattern recognition on graphs.

The paper is structured as follows: Section 2 introduces attributed graphs, the graph edit distance and graph edit kernels. In Section 3, we present the Graph Representation Theorem and derive general geometric results. Then in Section 4, we study the geometry of graph edit kernel spaces from the point of view of generic graphs. Finally, Section 5 concludes with a summary of the main result and indicates how the derived geometric results can be used in statistical pattern recognition.

\section{Graph Edit Kernel Spaces}
This section first introduces attributed graphs and the graph edit distance. To obtain graph spaces with a richer mathematical structure, we introduce graph metrics that are locally induced by inner products. We show that the derived graph metrics is (1) a special subclass of the graph edit distance, and (2) a common and widely used graph dissimilarity measure. For this purpose, we use a different formalization than presented in the literature. 

\subsection{Attributed Graphs}

Let $\S{A}$ be the set of node and edge attributes.  We assume that $\S{A}$ contains two (not necessarily distinct) symbols $N_{\S{V}}$ and $N_{\S{E}}$ denoting the null element for nodes and edges, respectively.

\begin{definition}
An attributed graph is a triple $X = (\S{V}, \S{E}, \alpha)$, where $\S{V}$ represents a finite set of nodes, $\S{E} \subseteq \S{V} \times \S{V}$ a set of edges, and $\alpha: \S{V} \times \S{V} \rightarrow \S{A}$ is an attribute function satisfying
\begin{enumerate}
\item $\alpha(i, j) \in \S{A}\setminus\cbrace{N_{\S{V}}, N_{\S{E}}}$, if $i \neq j$ and $(i, j) \in \S{E}$ 
\item $\alpha(i, j) = N_{\S{E}}$ if $i \neq j$ and $(i, j) \notin \S{E}$ 
\end{enumerate}
for all $i, j \in \S{V}$.
\end{definition}
The definition of an attributed graph implicitly assumes that graphs are fully connected by regarding non-edges as edges with null attribute $N_{\S{E}}$. Observe that nodes may have any attribute from $\S{A}$.  The node set of a graph $X$ is referred to as $\S{V}_X$, its edge set as $\S{E}_X$, and its attribute function as $\alpha_X$.  By $\S{G_A}$ we denote the set of attributed graphs with attributes from $\S{A}$.

Graphs can be directed and undirected. Attributes for node and edges may come from the same as well as from different or disjoint sets. For the sake of simplicity, we merged node and edge attributes into a single attribute set. Attributes can take any value. Examples are binary attributes, discrete attributes (symbols), continuous attributed (weights), vector-valued attributes, string attributes, and combinations thereof. Thus, the definition of attributed graphs is sufficient general to cover a wide class of graphs such as binary graphs from graph theory, weighted graphs, molecular graphs, protein structures, and many more.

\begin{definition}
A graph $Z$ is a subgraph of $X$, if
\begin{enumerate}
\item
 $\S{V}_Z \subseteq \S{V}_X$
 \item
 $\S{E}_Z \subseteq \S{E}_X$
 \item 
 $\alpha_Z = \alpha_X|_{\S{V}_Z \times \S{V}_Z}$.
 \end{enumerate}
\end{definition}

Suppose that $X$ and $Y$ are two graphs with $m$ and $n$ nodes, respectively. We say $X$ and $Y$ are size-aligned if both graphs are expanded to size $n+m$ by adding isolated nodes with attribute $N_{\S{V}}$. By $\widetilde{X}$ and $\widetilde{Y}$ we denote the size-aligned graphs of $X$ and $Y$. 

\begin{definition}
A morphism $\phi:X \rightarrow Y$ between graphs $X$ and $Y$ is a bijective map 
\[
\phi:\S{V}_{\widetilde{X}} \rightarrow \S{V}_{\widetilde{Y}}
\] 
between the node sets $\S{V}_{\widetilde{X}}$ and  $\S{V}_{\widetilde{Y}}$ of  the size-aligned graphs $\widetilde{X}$ and $\widetilde{Y}$, respectively. 
\end{definition}
Note that we use that same notation for morphism and its bijective node map. By $\S{M}_{X, Y}$ we denote the set of all morphisms from $X$ to $Y$.

\begin{definition}
An isomorphism is a morphism $\phi:X \rightarrow Y$ between graphs $X$ and $Y$ such that
\begin{align*}
\alpha_{\widetilde{X}}(i,j) = \alpha_{\widetilde{Y}}(\phi(i), \phi(j)) 
\end{align*}
for all $i, j \in \S{V}_{\widetilde{X}}$.
\end{definition}

Two graphs $X$ and $Y$ are isomorphic if and only if there is an isomorphism $\phi: X \rightarrow Y$ such that the restriction of $\phi$ to the unaligned node set $\S{V}_X$ satisfies
\[
\alpha_{X}(i,j) = \alpha_{Y}(\phi(i), \phi(j)) 
\]
for all $i, j \in \S{V}_X$. Thus the definition of isomorphism corresponds to the common definition of isomorphism from graph theory.

\subsection{Graph Edit Distance}

Next, we endow the set $\S{G_A}$ with a graph edit distance function. The basic idea of the graph edit distance is to regard a morphism $\phi : X \rightarrow Y$ as a transformation of a graph $X$ to a graph $Y$ by successively applying edit operations. Possible edit operations are insertion, deletion, and substitution of nodes and edges. Each node and edge edit operation is associated with a cost given by an edit cost function $\varepsilon:\S{A} \times \S{A} \rightarrow \R$.  Then the cost of transforming $X$ to $Y$ along morphism $\phi$ is the sum of the underlying edit costs. Table \ref{tab:edit-cost} provides an overview of different edit operations and the form of their edit costs. 

Let $\varepsilon:\S{A} \times \S{A} \rightarrow \R$ be an edit cost function. The cost of transforming $X$ to $Y$ along morphism $\phi: X \rightarrow Y$ is given by
\[
\delta_\phi(X, Y) = \sum_{i, j \in \S{V}_{\widetilde{X}}} \varepsilon\Big(\alpha_{\widetilde{X}}\big(i,j), \, \alpha_{\widetilde{Y}}\big(\phi(i), \phi(j)\big)\Big).
\]
The graph edit distance of $X$ and $Y$ minimizes the transformation cost over all possible morphisms between $X$ and $Y$.
\begin{definition}\label{def:ged}
Let $\varepsilon:\S{A} \times \S{A} \rightarrow \R$ be an edit cost function. The graph edit distance is a function $\delta: \S{G_A} \times \S{G_A} \rightarrow \R$ with 
\[
\delta(X, Y) = \min_{\phi \in \S{M}_{X, Y}} \delta_\phi(X, Y).
\]
\end{definition}
A graph edit distance space is a pair $\args{\S{G_A}, \delta}$ consisting of a set of attributed graphs together with a graph edit distance.

\begin{table}[t]
\centering
\begin{tabular}{ll}
\hline
\hline
cost & meaning\\
\hline
\\[-2ex]
$\varepsilon\args{N_{\S{V}}, y_{jj}}$ & insertion of node $y_{jj}$\\
$\varepsilon\args{x_{ii}, N_{\S{V}}}$ & deletion of node $x_{ii}$\\
$\varepsilon\args{x_{ii}, y_{jj}}$ & substitution of node $x_{ii}$ by node $y_{jj}$\\
$\varepsilon\args{N_{\S{V}}, N_{\S{V}}}$ & dummy operation with cost zero \\[1ex]
$\varepsilon\args{N_{\S{E}}, y_{rs}}$ & insertion of edge $y_{rs}$\\
$\varepsilon\args{x_{ij}, N_{\S{E}}}$ & deletion of edge $x_{ij}$\\
$\varepsilon\args{x_{ij}, y_{rs}}$ & substitution of edge $x_{ij}$ by edge $y_{rs}$\\
$\varepsilon\args{N_{\S{E}}, N_{\S{E}}}$ & dummy operation with cost zero \\
\hline
\hline
\end{tabular}
\caption{Overview of cost for edit operations. By $x_{ii}, y_{jj}$ we denote node attributes and by $x_{ij}, y_{rs}$ edge attributes. Costs $c\args{N_{\S{V}}, N_{\S{V}}}$ and $c\args{N_{\S{E}}, N_{\S{E}}}$ are zero and arise by expansion of graphs for mathematical convenience.}
\label{tab:edit-cost}
\end{table}

\subsection{Graph Edit Kernels}

Graph spaces endowed with the graph edit distance are difficult to analyze. To obtain spaces that are mathematically more structured, we impose constraints on the set of feasible morphisms and the choice of edit costs. 

\medskip

First, we constrain the set of morphisms to the subset of compact morphisms.
\begin{definition}
A morphism $\phi:X \rightarrow Y$ between graphs $X$ and $Y$ is compact, if 
\[
\phi(\S{V}_X) \subseteq \S{V}_Y \quad \text{or} \quad  \phi^{-1}(\S{V}_Y) \subseteq \S{V}_X.
\] 
By $\S{C}_{X,Y}$ we denote the subset of compact morphisms between $X$ and $Y$.
\end{definition}

A compact morphism demands that each node of the smaller of both graphs corresponds to a unique node of the larger one. 

Next, we constrain the choice of edit cost via edit scores for measuring the similarity of node and edge attributes. We consider edit score functions of the form
\[
k :\S{A} \times \S{A} \rightarrow \R, \quad (x, y) \mapsto  \Phi(x)^T\Phi(y),
\]
where $\Phi: \S{A} \rightarrow \S{H}$ is a feature map into a Hilbert space $\S{H}$. Then the edit score is a positive definite kernel on $\S{A}$. 
The score of transforming $X$ to $Y$ along a compact morphism $\phi: X \rightarrow Y$ is given by
\[
\kappa_\phi(X, Y) = \sum_{i, j \in \S{V}_{\widetilde{X}}} k\Big(\alpha_{\widetilde{X}}\big(i,j), \, \alpha_{\widetilde{Y}}\big(\phi(i), \phi(j)\big)\Big).
\]
Maximizing the transformation score over all compact morphisms gives the graph edit kernel. 
\begin{definition}
The graph edit kernel is a function $\kappa: \S{G_A} \times \S{G_A} \rightarrow \R$ with 
\[
\kappa(X, Y) = \max_{\phi \in \S{C}_{X, Y}} \kappa_{\phi}(X, Y).
\]
\end{definition}
As an optimal assignment kernel, the graph edit kernel is not positive definite \cite{Vert2008}, but gives rise to a metric.
\begin{proposition}\label{prop:gekd}
A graph edit kernel $\kappa$ induces a metric $\delta: \S{G_A} \times \S{G_A} \rightarrow \R$ defined by
\begin{align}\label{eq:gekd}
\delta(X,  Y) = \sqrt{\kappa(X, X) + \kappa(Y, Y) - 2\kappa(X, Y)}
\end{align}
for all $X, Y \in \S{G_A}$. 
\end{proposition}
\proof
Follows from Theorem \ref{theorem:geodesic-space}.
\qed

\medskip 

We call the graph edit distance $\delta$ defined in Prop.~\ref{prop:gekd} the metric induced by the graph edit kernel $\kappa$.  A graph edit kernel space is a graph edit distance space $(\S{G_A}, \delta)$, where $\delta$ is a metric induced by a graph edit kernel.

An equivalent way to derive the metric defined in  \eqref{eq:gekd} is as follows: suppose that $k(x, y) = \Phi(x)^T\Phi(y)$ is a positive definite kernel. Define the edit cost function
\begin{align*}
\varepsilon_k: \S{A} \times \S{A} \rightarrow \R, \quad (x, y) \mapsto \normS{\Phi(x)-\Phi(y)}{^2},
\end{align*}
where the norm $\norm{\cdot}$ in $\S{H}$ is induced by the inner product in the usual way. Applying the kernel-trick gives
\begin{align}\label{eq:kernel-trick}
\varepsilon_k(x, y) = k(x, x) + k(y, y) - 2 k(x, y).
\end{align}
Then the edit cost function $\varepsilon_k$ induces a graph edit distance that coincides with the squared metric defined in \eqref{eq:gekd}.

\subsection{Examples}

The following examples show that the graph edit kernel and its induced graph edit kernel distance are not artificial constructions, but comprise well known and widely applied structural (dis)similarity measures for graphs. 

\subsubsection{Maximum Common Subgraph}

The first example shows that the maximum common subgraph problem is equivalent to the problem of computing a graph edit kernel. For this, we first introduce two definitions.

\begin{definition}
A common subgraph of $X$ and $Y$ is a graph $Z$ that is isomorphic to subgraphs $Z_X$ of $X$ and $Z_Y$ of $Y$. 
\end{definition}
Let $\S{S}_{X,Y}$ denote the set of all common subgraphs of graphs $X$ and $Y$. Then a maximum common subgraphs of two graphs is a subgraph $Z^*$ with maximum number of nodes and edges.
\begin{definition}
A maximum common subgraph of $X$ and $Y$ is a common subgraph satisfying
\[
Z^* = \arg\max_{Z \in \S{S}_{X, Y}} \abs{\S{V}_Z} + \abs{\S{E}_Z}.
\]
\end{definition}

The function
\[
k :\S{A} \times \S{A} \rightarrow \R, \quad (x, y) \mapsto \begin{cases}
1 & x = y\\
0 & \text{otherwise}.
\end{cases}
\]
is a positive definite kernel that gives rise to a graph edit kernel
\[
\kappa(X, Y) = \max_{Z \in \S{S}_{X, Y}} \abs{\S{V}_Z} + \abs{\S{E}_Z}.
\]

\subsubsection{Geometric Graph Metrics}

Let $\S{A} = \R$ be the set of weights with null elements $N_{\S{V}} = N_{\S{E}} = 0$. Then $\S{G_A}$ is the set of (positively) weighted graphs. We can represent a graph $X$ with $n$ nodes by a weighted adjacency matrix $\vec{X} = (x_{ij})$ from $\R^{n \times n}$ with elements $x_{ij} = \alpha_X(i, j)$.

The function 
\begin{align}\label{eq:kernel:xy}
k :\S{A} \times \S{A} \rightarrow \R, \quad (x, y) \mapsto x \cdot y
\end{align}
is a positive definite kernel as an inner product of the one-dimensional vector space $\S{A}$. The induced edit cost function takes the form 
\[
\varepsilon(x, y) = \argsS{x-y}{^2}.
\]
Let $\vec{X} \in \R^{n \times n}$ and $\vec{Y} \in \S{R}^{m \times m}$ be weighted adjacency matrices of graphs $X$ and $Y$, respectively. Then the graph metric induced by kernel \eqref{eq:kernel:xy} is of the form
\[
\delta(X, Y) = \min_{\vec{P} \in \Pi}\norm{\vec{X} -\vec{P}\vec{Y}\vec{P}^T},
\]
where $\Pi$ denotes the set of all $(n \times m)$ subpermutation matrices of rank $d = \min(n, m)$. A subpermutation matrix of rank $d$ is a matrix that satisfies the following conditions:
\begin{enumerate}
\item All matrix elements are from $\cbrace{0,1}$.
\item Each row and each column has at most one element with value $1$.
\item The matrix has exactly $d$ elements with value $1$.
\end{enumerate}

Geometric graph metrics can be easily generalized to the case where node and edge attributes are from different Euclidean spaces, say $\R^p$ and $\R^q$.

\section{The Graph Representation Theorem}\label{sec:proofs}

The Graph Representation Theorem states that a graph of bounded order can be represented as a point in an orbit space. This result is useful, because it provides deep insight into the geometry of graph edit kernel spaces and simplifies derivation of many results. 

\paragraph*{Assumptions.} We assume that the underlying edit score $k:\S{A} \times \S{A} \rightarrow \R$ is defined as an inner product of the feature map $\Phi: \S{A} \rightarrow \S{H}$ into the $d$-dimensional Euclidean space $\S{H} = \R^d$. By $\S{G_H}$ we denote the space of attributed graphs  of all graphs of bounded order $n$ with attributes from the feature space $\S{H}$.\footnote{The order of a graph is the number of its nodes.} We may regard $\S{G_A}$ as a subset of $\S{G_H}$ via the feature map $\Phi$. By $\kappa: \S{G_H} \times \S{G_H} \rightarrow \R$ we denote the graph edit kernel based on edit score $k$, and by $\delta: \S{G_H} \times \S{G_H} \rightarrow \R$ the metric induced by the graph edit kernel $\kappa$. 

\subsection{The Graph Representation Theorem}
Let $\S{G}$ be a group with neutral element $\varepsilon$.  An action of group $\S{G}$ on a set $\S{X}$ is a map
\[
\phi: \S{G} \times \S{X} \rightarrow \S{X}, \quad (\gamma, x) \mapsto \gamma x
\]
satisfying
\begin{enumerate}
\item $(\gamma \circ \gamma') x = \gamma(\gamma' x)$
\item $\varepsilon x = x$
\end{enumerate}
for all $\gamma, \gamma' \in \S{G}$ and all $x \in \S{X}$. The orbit of $x \in \S{X}$ under the action $\S{G}$ is the subset of $\S{X}$ defined by
\[
\bracket{x} = \cbrace{\gamma x \,:\, \gamma \in \S{G}}.
\]
We write $x' \in [x]$ to denote that $x'$ is an element of the orbit $[x]$. The \emph{orbit space} of the action of $\S{G}$ on $\S{X}$ is defined to bet the set of all orbits
\[
\S{X}/\S{G} = \cbrace{[x] \,:\, x \in \S{X}}.
\]

The next result states that each graph can be represented as a points of an orbit space.
\begin{theorem}[Graph Representation Theorem]\label{theorem:grt}
A graph edit kernel space  $\args{\S{G_H}, \delta}$ is isometric to the orbit space $\S{X}/\S{G}$ of the action of a group $\S{G}$ of isometries on a Euclidean space $\S{X}$. 
\end{theorem}

\noindent 
\proof
We present a constructive proof. 

\begin{part}
First we show that we may assume that all graphs from $\S{G_H}$ are of order $n$ without loss of generality. 
Suppose that $X = \args{\S{V}, \S{E}, \alpha}$ is a graph. An isolated node $i \in \S{V}$ is a node without connection to any other node, that is
\[
\alpha(i, j) = \alpha(j,i) = N_{\S{E}}
\]
for all $j \in \S{V} \setminus \cbrace{i}$, where $N_{\S{E}}$ is the null attribute for denoting non-existence of an edge. An isolated node $i \in \S{V}$ is a null-node if $\alpha(i, i) = N_{\S{V}}$, where $N_{\S{V}}$ is the null attribute for nodes. Suppose that $X'$ is a graph obtained by removing or adding null-nodes. Then by definition, the graphs $X$ and $X'$ are isomorphic. Thus, if $X$ is of order $m< n$, we replace $X$ by a graph $X'$ of order $n$ by augmenting $X$ with $n-m$ null-nodes.
\end{part}

\begin{part}
Let $\S{X} = \S{H}^{n \times n}$ be the set of all $(n \times n)$-matrices with elements from $\S{H}$. An attributed graph $X= \args{\S{V}, \S{E}, \alpha}$ is completely specified by a matrix $\vec{X} = \args{\vec{x}_{ij}}$ from $\S{X}$ with elements $x_{ij} = \alpha(i,j)$ for all $i, j \in \S{V}$. 
\end{part}

\begin{part}
The form of matrix $\vec{X}$ is generally not unique and depends on how the nodes are arranged in the diagonal of $\vec{X}$. Different orderings of the nodes may result in different matrix representations. The set of all possible re-orderings of all nodes of $X$ is (isomorphic to) the symmetric group $\S{S}_n$, which in turn is isomorphic to the group $\S{G}$ of all simultaneous row and column permutations of a matrix from $\S{X}$. Thus, we have a group action
\[
\S{G} \times \S{X} \rightarrow \S{X}, \quad (\gamma, \vec{X}) \mapsto \gamma \vec{X}, 
\]
where $\gamma \vec{X}$ denotes the matrix obtained by simultaneously permuting the rows and columns according to $\gamma$. For $\vec{X} \in \S{X}$, the orbit of $\vec{X}$ is the set defined by
\[
\bracket{\vec{X}} = \cbrace{\gamma \vec{X} \,:\, \gamma \in \S{G}}.
\]	
By
\[
\S{X}/\S{G} = \cbrace{\bracket{\vec{X}} \,:\, \vec{X} \in \S{X}} 
\] 
we denote the orbit space consisting of all all orbits. The natural projection map is defined by 
\[
\pi: \S{X} \rightarrow \S{X}/\S{G}, \quad \vec{X} \mapsto \bracket{\vec{X}}.
\]
\end{part}

\begin{part} 
To emphasize that $\S{X}$ is a Euclidean space, we use vector instead of matrix notation henceforth. Consequently, we write $\vec{x}$ instead of $\vec{X}$. By $[\vec{x}]$ we denote the orbit of $\vec{x}$. 
\end{part}

\begin{part}\label{part:quotient_topology}
The quotient topology on $\S{X}/\S{G}$ is the finest topology for which the projection map $\pi$ is continuous. Thus, the open sets $\S{U}$ of the quotient topology on $\S{X}/\S{G}$ are those sets for which $\pi^{-1}(\S{U})$ is open in $\S{X}$.
\end{part}

\begin{part}\label{part:quotient_distance}
We can endow $\S{X}/\S{G}$ with the quotient distance defined by
\[
d\!\args{\bracket{\vec{x}}, \bracket{\vec{y}}} = \inf \; \sum_{i=1}^k \norm{\vec{x}_i - \vec{y}_i},
\]
where the infimum is taken over all finite sequences $\args{\vec{x}_1, \ldots. \vec{x}_k}$ and $\args{\vec{y}_1, \ldots. \vec{y}_k}$ with $\bracket{\vec{x}_1} = \bracket{\vec{x}}$, $\bracket{\vec{y}_n} = \bracket{\vec{y}}$, and $\bracket{\vec{x}_i} = \bracket{\vec{y}_{i+1}}$ for all $i \in \cbrace{1, \ldots, k-1}$. The distance $d\args{\bracket{\vec{x}}, \bracket{\vec{y}}} $ is a pseudo-metric \cite{Bridson1999}, Lemma 5.20. Since $\S{G}$ is finite and acts by isometries, the pseudo-metric is a metric of the form 
\begin{align*}
d\!\args{\bracket{\vec{x}}, \bracket{\vec{y}}} &= \min \cbrace{\norm{\vec{x}-\vec{y}} \,:\, \vec{x} \in \bracket{\vec{x}}, \vec{y} \in \bracket{\vec{y}}}\\
&= \min \cbrace{\norm{\vec{x}-\vec{y}} \,:\, \vec{x} \in \bracket{\vec{x}}}\\
&= \min \cbrace{\norm{\vec{x}-\vec{y}} \,:\, \vec{y} \in \bracket{\vec{y}}},
\end{align*}
where $\vec{y}$ in the second row and $\vec{x}$ in the third row are arbitrarily chosen representations. 
\end{part}

\begin{part}
The quotient metric $d\!\args{\bracket{\vec{x}}, \bracket{\vec{y}}}$ on $\S{X}/\S{G}$  defined in part \ref{part:quotient_distance} induces a topology that coincides with the quotient topology defined in part \ref{part:quotient_topology}.
\end{part}

\begin{part}\label{part:isometric-isomorphism}
Consider the map 
\[
\omega: \S{G_H} \rightarrow \S{X}/\S{G}, \quad X \mapsto \bracket{\vec{x}} 
\]
that assigns each graph $X$ to the orbit $[\vec{x}]$ consisting of all representations of $X$. The map $\omega$ is surjective, because any matrix $\vec{x} \in \S{X}$ represents a valid graph $X$. The problem of dangling edges is solved by allowing nodes with null attribute. Then the orbit $\bracket{X}$ consists of all matrices representing $X$. The map $\omega$ is also  injective. Suppose that $X$ and $Y$ are non-isomorphic graphs with respective representations $\vec{x}$ and $\vec{y}$. If $\vec{x}$ and $\vec{y}$ are in the same orbit, then there is an isomorphism between $X$ and $Y$, which contradicts our assumption. Thus, $\omega$ is a bijection.
\end{part}

\begin{part}\label{part:isometry} We have
\begin{align*}
\delta^2(X, Y) &= \min_{\phi \in \S{C}_{X,Y}} \;\sum_{i, j \in \S{V}_X} \normS{\alpha_X(i,j) - \alpha_Y(\phi(i), \phi(j))}{^2}\\
&= \min_{\phi \in \S{C}_{X,Y}} \;\sum_{i, j \in \S{V}_X}  \normS{\vec{x}_{ij} - \vec{y}_{\phi(i)\phi(j)}}{^2}\\
&= \min_{\gamma \in \S{G}} \normS{\vec{x} - \gamma(\vec{y})}{^2}\\
&= d^2(\bracket{\vec{x}}, \bracket{\vec{y}})
\end{align*}
for all $X, Y \in \S{G_H}$. The first equation follows from the definition of a graph edit distance induced by a graph edit kernel. The second equation changes the notation of the attributes. The third equation follows from the equivalence of morphism and group action by construction. Finally, the fourth equation follows from part \ref{part:quotient_topology}.
\end{part}

\begin{part}
The implication of part \ref{part:isometry} are twofold:
\begin{enumerate}
\item The distance $\delta(X, X)$ induced by the graph edit kernel $\kappa$ is a metric.
\item The map $\omega: \S{G_H} \rightarrow \S{X}/\S{G}$ defined in part \ref{part:isometric-isomorphism} is an isometry.
\end{enumerate}
This shows the assertion of the theorem.
\end{part}
\qed

\bigskip

Corollary \ref{Corollary:GRT} summarizes results obtained in the course of proving the Graph Representation Theorem. 
\begin{corollary}\label{Corollary:GRT}
\ 
\begin{enumerate}
\item 
The natural projection $\pi: \S{X} \rightarrow \S{X}/\S{G}$ is continuous.
\item 
The quotient distance $d$ on $\S{X}/\S{G}$ is a metric.
\item The map 
\[
\omega: \S{G_H} \rightarrow \S{X}/\S{G}, \quad X \mapsto \bracket{\vec{x}}
\]
is a bijective isometry between the metric spaces $\args{\S{G_H}, \delta}$ and $\args{\S{X}/\S{G}, d}$.
\item 
The distance $\delta$ on $\S{G_H}$ induced by the graph edit kernel $\kappa$ is a metric satisfying
\begin{align*}
\delta(X, Y) &= \min \cbrace{\norm{\vec{x} - \vec{y}}\,:\, \vec{x} \in \omega(X), \vec{y} \in \omega(Y)}\\
&= \min \cbrace{\norm{\vec{x} - \vec{y}}\,:\, \vec{x} \in \omega(X)} \; \text{ for all } \;\vec{y}\in \omega(Y)\\
&= \min \cbrace{\norm{\vec{x} -\vec{y}}\,:\, \vec{y} \in \omega(Y)}  \; \text{ for all } \;\vec{x}\in \omega(X)
\end{align*}
for all $X, Y \in \S{G_H}$. 
\item 
The graph edit kernel $\kappa$ satisfies
\begin{align*}
\kappa(X, Y) &= \max \cbrace{\vec{x}^T\vec{y}\,:\, \vec{x} \in \omega(X), \vec{y} \in \omega(Y)}\\
&= \max \cbrace{\vec{x}^T\vec{y}\,:\, \vec{x} \in \omega(X)}  \; \text{ for all } \;\vec{y}\in \omega(Y)\\
&= \max \cbrace{\vec{x}^T\vec{y}\,:\, \vec{y} \in \omega(Y)} \; \text{ for all } \;\vec{x}\in \omega(X)
\end{align*}
for all $X, Y \in \S{G_H}$.
\end{enumerate}
\end{corollary}

\bigskip

Studying graph edit kernel spaces reduces to the study of orbit spaces $\S{X}/\S{G}$ due to the Graph Representation Theorem. Though analysis of orbit spaces is more general, we translate all results into graph edit kernel spaces to make them directly accessible for statistical pattern recognition methods. For this, we identify the graph space $\S{G_H}$ with the orbit space $\S{X}/\S{G}$ via the bijective isometry $\omega: \S{G_H} \rightarrow \S{X}/\S{G}$. We denote this relationship by $\S{G_H} \cong \S{X}/\S{G}$. Suppose that $X \in \S{G_H}$ and $\vec{x} \in \S{X}$. We briefly write $\vec{x} \in X$ if $\omega(X) = \bracket{\vec{x}}$. In this case, we call $\vec{x}$ a \emph{representation} of graph $X$.

\medskip

Next, we summarize some results useful for a statistically consistent analysis of graphs. 
\begin{theorem}\label{theorem:geodesic-space}
A graph edit kernel space  $\args{\S{G_H}, \delta}$ has the following properties:
\begin{enumerate}
\item $\args{\S{G_H}, \delta}$ is a complete metric space.
\item $\args{\S{G_H}, \delta}$ is a geodesic space.
\item $\args{\S{G_H}, \delta}$ is locally compact.
\item Every closed bounded subset of $\args{\S{G_H}, \delta}$ is compact.
\end{enumerate}
\end{theorem}

\noindent
\proof
Since $\S{G_H} \cong \S{X}/\S{G}$ it is sufficient to show the assertions for the orbit space  $\S{X}/\S{G}$.
\setcounter{part_counter}{0}
\begin{part}
Since the group $\S{G}$ is finite, all orbits are finite and therefore closed subsets of $\S{X}$. The Euclidean space $\S{X}$ is a finitely compact metric space. Then $\S{X}/\S{G}$ is a complete metric space \cite{Ratcliffe2006}, Theorem 8.5.2. 
\end{part}
\begin{part}
Since $\S{X}$ is a finitely compact metric space and $\S{G}$ is a  discontinuous group of isometries, the assertion follows from \cite{Ratcliffe2006}, Theorem 13.1.5. 
\end{part}
\begin{part}
Since $\S{G}$ is finite and therefore compact group, the assertion follows from \cite{Bredon1972}, Theorem 3.1.
\end{part}
\begin{part}
Since $\S{G_H}$ is a complete, locally compact length space, the assertion follows from the Hopf-Rinow Theorem (see e.g.\cite{Bridson1999}. Prop.~3.7).
\end{part}
\qed

\subsection{Length and Angle}
In this section, we introduce basic geometric concepts such as length and angle of graphs, which are important for a geometric interpretation and understanding of generalized linear methods for classification of graphs \cite{Jain2014a,Jain2014b}.

\begin{definition}
The scalar multiplication on $\S{G_H}$ is a function 
\[
\cdot: \R \times \S{G_H} \rightarrow \S{G_H}, \quad (\lambda, X) \mapsto  \lambda X,
\]
where $\lambda X$ is the graph obtained by scalar multiplication of $\lambda$ with all node and edge attributes of $X$. 
\end{definition}

In contrast to scalar multiplication on vectors, scalar multiplication on graphs is only positively homogeneous.
\begin{proposition}\label{prop:positively-homogeneous}
Let $\lambda \in \R_+$ a non-negative scalar. Then we have
\[
\kappa(X, \lambda Y) = \lambda \kappa(X, Y)
\] 
for all $X, Y \in \S{G_H}$.
\end{proposition}
\noindent
\proof
From $\S{G_H} \cong \S{X}/\S{G}$ and Corollary \ref{Corollary:GRT} follows
\[
\kappa(X, Y) = \max \cbrace{\vec{x}^T\vec{y} \,:\, \vec{x} \in X},
\]
where $\vec{y} \in Y$ is arbitrary but fixed. Suppose that $\kappa(X, Y) = \vec{x}_0^T\vec{y}$ for some representation $\vec{x}_0 \in X$. Then we have
$\vec{x}_0^T\vec{y} \geq \vec{x}^T\vec{y}$ for all $\vec{x} \in X$. This implies
\begin{align*}
\vec{x}_0^T(\lambda\vec{y}) = \lambda\args{\vec{x}_0^T\vec{y}}  \geq  \lambda\args{\vec{x}^T\vec{y}} = \vec{x}^T(\lambda\vec{y})
\end{align*}
for all non-negative scalars $\lambda \in \R_+$.  The assertion follows from $\lambda Y = \bracket{\lambda \vec{y}}$ if  $Y= \bracket{\vec{y}}$. 
\qed

\medskip

Using graph edit kernels, we can define the length of a graph in the usual way.
\begin{definition}
The length $\ell(X)$ of graph $X \in \S{G_H}$ is defined by
\[
\ell(X) = \sqrt{\kappa(X, X)}.
\]
\end{definition}
The length of a graph can be determined efficiently, because the transformation score of the identity morphism is maximum over all morphisms from a graph to itself. 
\begin{proposition}\label{prop:length}
The squared length of $X$ is of the form
\[
\ell(X) = \norm{\vec{x}} = \sqrt{\kappa_{\id}(X, X)} = \sqrt{ \sum_{i, j \in \S{V}_X} k(i, j)} 
\]
for all $\vec{x} \in X$.
\end{proposition}

\noindent
\proof
From Corollary \ref{Corollary:GRT} follows
\[
\kappa(X, X) = \max \cbrace{\vec{x}^T\vec{x}' \,:\, \vec{x},\vec{x}' \in X}.
\]
We have 
\[
\vec{x}^T\vec{x}' = \norm{\vec{x}} \norm{\vec{x}'} \cos \alpha,
\]
where $\alpha$ is the angle between $\vec{x}$ and $\vec{x}'$. Since $\S{G}$ is a group of isometries acting on $\S{X}$, we have 
\[
\norm{\vec{x}} = \norm{\vec{x}'}
\]
for all elements $\vec{x}$ and $\vec{x}'$ from the same orbit.  Thus, $\vec{x}^T\vec{x}' $ is maximum if the angle $\alpha \in [0, 2\pi]$ is minimum over all pairs of elements from $X$. The minimum angle is zero for pairs of identical elements. This shows the assertion.
\qed

\medskip

The relationship between the length of a graph and the graph edit kernel is given by a weak form of the Cauchy Schwarz inequality:
\begin{theorem}[Weak Cauchy-Schwarz]\label{prop:proved-cauchy-schwarz}
Let $X, Y \in \S{G_H}$ be two graphs. Then we have
\[
\abs{\kappa(X, Y)} \leq \ell(X)\cdot\ell(Y),
\] 
where equality holds when $X$ and $Y$ are positively dependent.
\end{theorem}
\noindent
\proof
From $\S{G_H} \cong \S{X}/\S{G}$ and Corollary \ref{Corollary:GRT} follows
\[
\kappa(X, Y) = \max \cbrace{\vec{x}^T\vec{y} \,:\, \vec{x} \in X, \vec{y} \in Y}
\]
for all $X, Y \in \S{G_H}$. Suppose that $\vec{x}_0 \in X$ and $\vec{y}_0 \in Y$ are representations such that $\kappa(X, Y) = \vec{x}_0^T\vec{y}_0$. From the standard Cauchy-Schwarz inequality together with Prop.~\ref{prop:length} follows
\[
\abs{\kappa(X, Y)} = \abs{\vec{x}_0^T\vec{y}_0} \leq \norm{\vec{x}_0} \norm{\vec{y}_0} = \ell(X) \ell(Y).
\]
We show the second assertion, that is equality if $X$ and $Y$ are positively dependent. Suppose that $X = \lambda Y$ for some $\lambda \in \R_+$. From the definition of the length of a graph together with Prop.~\ref{prop:positively-homogeneous} follows
\[
\lambda^2\ell^2(X) = \lambda^2 \kappa(X, X) = \kappa(\lambda X, \lambda X) = \ell^2(\lambda X).
\]
Then by using Prop.~\ref{prop:positively-homogeneous} we obtain
\[
\abs{\kappa(X, \lambda X) } = \lambda \kappa(X, X) = \lambda \ell(X) \ell(X) =  \ell(X) \ell(\lambda X).
\]
\qed

\medskip

The Cauchy-Schwarz inequality is considered to be weak, because equality holds only for positively dependent graphs. This is in contrast to the original Cauchy-Schwarz inequality in vector spaces, where equality holds, when two vectors are linearly dependent. Nevertheless, we can use the weak Cauchy-Schwarz inequality for defining an angle between two graphs.

\begin{definition}
The cosine of the angle between non-zero graphs $X$ and $Y$ is defined by
\begin{align}\label{eq:angle}
\cos \alpha = \frac{\kappa(X, Y)}{\ell(X)\ell(Y)}.
\end{align}
\end{definition}
With the notion of angle, we can introduce orthogonality between graphs. 

\begin{definition}
Two graphs $X$ and $Y$ are orthogonal, if $\kappa(X, Y) = 0$. A graph $X$ is orthogonal to a subset $\S{U} \subseteq \S{G_H}$, if 
\[
\kappa(X,Y) - \kappa(X, Z) = 0
\]
for all $Y, Z \in \S{U}$.
\end{definition}

\subsection{Geometry from a Generic Viewpoint}

In this section, we describe the geometry of a graph edit kernel space from a generic viewpoint. A generic property is defined as follows:
\begin{definition}
A generic property is a property that holds on a dense open set. 
\end{definition}
Suppose that the metric space $(\S{M}, d)$ is either a Euclidean space or graph edit kernel space. 
The underlying topology that determines the open subsets of $\S{M}$ is the topology induced by the metric $d$. The open sets of the topology are all subsets that can be realized as the unions of open balls
\[
\S{B}(z, \rho) = \cbrace{x \in \S{M} \,:\, d(z, x) < \rho},
\]
with center $z\in \S{M}$ and radius $\rho > 0$. In measure-theoretic terms, a generic property is a property that holds \emph{almost everywhere}, meaning for all points of a set with Borel probability measure one.

\subsubsection{Dirichlet Fundamental Domains}
We assume that $\S{G}$ is non-trivial. In the trivial case, we have $\S{X} = \S{X}/\S{G}$ and therefore everything reduces to the geometry of Euclidean spaces. For every $\vec{x} \in \S{X}$, we define the isotropy group of $\vec{x}$ as the set
\[
\S{G}_{\vec{x}} = \cbrace{\gamma \in \S{G} \,:\, \gamma \vec{x} = \vec{x}}.
\]
An ordinary point $\vec{x} \in X$ is a point with trivial isotropy group $\S{G}_{\vec{x}} = \cbrace{\varepsilon}$. A singular point is a point with non-trivial isotropy group. If $\vec{x}$ is ordinary, then all elements of the orbit $\bracket{\vec{x}}$ are ordinary points. 

A subset $\S{F}$ of $\S{X}$ is a \emph{fundamental set} for $\S{G}$ if and only if $\S{F}$ contains exactly one point $\vec{x}$ from each orbit $\bracket{\vec{x}} \in \S{X}/\S{G}$. 
A \emph{fundamental domain} of $\S{G}$ in $\S{X}$ is a closed set $\S{D} \subseteq \S{X}$ that satisfies 
\begin{enumerate}
\item $\S{X} = \bigcup_{\gamma \in \S{G}} \gamma\S{D}$
\item $\gamma \S{D}^\circ \cap \S{D}^\circ = \emptyset$ for all $\gamma \in \S{G} \setminus \cbrace{\varepsilon}$.
\end{enumerate}

\begin{proposition}
Let $\vec{z}\in \S{X}$ be ordinary. Then the set 
\[
\S{D}_{\vec{z}} = \cbrace{\vec{x} \in \S{X} \,:\, \norm{\vec{x} - \vec{z}} \leq \norm{\vec{x} - \gamma \vec{z}} \text{ for all } \gamma \in \S{G}}
\]
is a fundamental domain, called \emph{Dirichlet fundamental domain} centered at $\vec{z}$. 
\end{proposition}
\noindent
\proof \cite{Ratcliffe2006}, Theorem 6.6.13. \qed

\medskip

\begin{proposition}\label{proof:properties-of-DFD}
Let $\S{D}_{\vec{z}}$ be a Dirichlet fundamental domain centered at an ordinary point $\vec{z}$. Then the following properties hold:
\begin{enumerate}
\item $\S{D}_{\vec{z}}$ is a convex polyhedral cone.
\item There is a fundamental set $\S{F}_{\vec{z}}$ such that 
\[
\S{D}_{\vec{z}}^\circ \subseteq \S{F}_{\vec{z}} \subseteq \S{D}_{\vec{z}}.
\]
\item We have $\vec{z}  \in \S{D}_{\vec{z}}^\circ$.
\item Every point $\vec{x} \in \S{D}_{\vec{z}}^\circ$ is ordinary.
\item Suppose that $\vec{x}, \gamma \vec{x} \in \S{D}_{\vec{z}}$ for some  $\gamma \in \S{G}\setminus\cbrace{\varepsilon}$. Then $\vec{x}, \gamma \vec{x} \in \partial \S{D}_\mu$.
\item The Dirichlet fundamental domain can be equivalently expressed as 
\[
\S{D}_{\vec{z}} = \cbrace{\vec{x} \in \S{X} \,:\, \vec{x}^T\vec{z} \geq \vec{x}^T\gamma \vec{z} \text{ for all } \gamma \in \S{G}}.
\]
\end{enumerate}
\end{proposition}

\noindent
\proof
\setcounter{part_counter}{0}
\begin{part}
For each $\gamma \neq \varepsilon$, we define the closed halfspace $\S{H}_\gamma = \cbrace{\vec{x} \in \S{X} \,:\, \norm{\vec{x} - \vec{z}} \leq \norm{\vec{x} - \gamma\vec{z}}}$. Then 
the Dirichlet fundamental domain $\S{D}_{\vec{z}}$ is of the form
\[
\S{D}_{\vec{z}} = \bigcap_{\gamma \in \S{G}} \S{H}_\gamma.
\]
As an intersection of finitely many closed halfspaces, the set  $\S{D}_{\vec{z}}$  is a convex polyhedral cone \cite{Gale1951}. 
\end{part}

\begin{part}
\cite{Ratcliffe2006}, Theorem 6.6.11.
\end{part}

\begin{part}
The isotropy group of an ordinary point is trivial. Thus  $\vec{z}^T\vec{z} > \vec{z}^T\gamma\vec{z}$ for all $\gamma \in \S{G} \setminus \cbrace{\varepsilon}$. This shows that $\vec{z}$ lies in the interior of $\S{D}_{\vec{z}}$.
\end{part}

\begin{part}
Suppose that  $\vec{x} \in \S{D}_{\vec{z}}^\circ$ is singular. Then the isotropy group $\S{G}_{\vec{x}}$ is non-trivial. Thus, there is a $\gamma \in \S{G}\setminus\cbrace{\varepsilon}$ with $\vec{x} = \gamma \vec{x}$. This implies $\vec{x} \in \gamma \S{D}_{\vec{z}} \cap  \S{D}_{\vec{z}}$. Then $\vec{x} \in \partial\S{D}_{\vec{z}}$ is a boundary point of $\S{D}_{\vec{z}}$ by   \cite{Ratcliffe2006}, Theorem 6.6.4. This contradicts our assumption $\vec{x} \in \S{D}_{\vec{z}}^\circ$ and shows that $\vec{x}$ is ordinary.
\end{part}

\begin{part}
From $\vec{x}, \gamma \vec{x} \in \S{D}_{\vec{z}}$ follows $\norm{\vec{x} - \vec{z}} = \norm{\gamma\vec{x} - \vec{z}}$. Since $\S{G}$ acts by isometries, we have  $\norm{\vec{x} - \vec{z}} = \norm{\gamma\vec{x} - \gamma\vec{z}}$. Combining both equations yields $\norm{\gamma\vec{x} - \vec{z}} = \norm{\gamma\vec{x} - \gamma\vec{z}}$. This shows that $\gamma\vec{x} \in \partial \S{D}_\mu$. Let $\gamma' \in \S{G}$ be the inverse of $\gamma$. Since $\gamma \neq \varepsilon$, we have $\gamma' \neq \varepsilon$. Then
\begin{align*}
\norm{\vec{x} - \vec{z}} &=\norm{\gamma\vec{x} - \vec{z}} 
= \norm{\gamma'\gamma\vec{x} - \gamma'\vec{z}}
= \norm{\vec{x} - \gamma'\vec{z}},
\end{align*}
where the second equation follows from isometry of the group action. Thus, $\norm{\vec{x} - \vec{z}} = \norm{\vec{x} - \gamma'\vec{z}}$ shows that $\vec{x} \in \partial \S{D}_\mu$.
\end{part}

\begin{part}
The following equivalences hold for all $\gamma \in \S{G}$:
\begin{align*}
& \vec{x} \in \S{D}_{\vec{z}}\\
\Leftrightarrow  \quad & \normS{\vec{x} - \vec{z}}{^2} \leq  \normS{\vec{x} - \gamma\vec{z}}{^2}\\
\Leftrightarrow  \quad & \normS{\vec{x}}{^2} + \normS{\vec{z}}{^2} - 2\vec{x}^T\vec{z} \leq \normS{\vec{x}}{^2} + \normS{\gamma\vec{z}}{^2} - 2\vec{x}^T\gamma\vec{z}\\
\Leftrightarrow  \quad & \vec{x}^T\vec{z} \geq \vec{x}^T\gamma\vec{z}.
\end{align*}
The last equivalence uses that $\S{G}$ acts on $\S{X}$ by isometries. This shows the last property.
\end{part}
\qed

\medskip

\begin{corollary}\label{cor:generic-ordinary}
A generic point $\vec{z} \in \S{X}$ is ordinary.
\end{corollary}

\noindent
\proof
Suppose that $\vec{z} \in \S{X}$ is ordinary. Then there is a Dirichlet fundamental domain $\S{D}_{z}$. From Prop.~\ref{proof:properties-of-DFD} follows that all points of the open set $\S{D}_{z}^\circ$ are ordinary. With $\vec{z}$ all representatives from $\bracket{\vec{z}}$ are ordinary. Then all points of $\gamma\S{D}_{z}^\circ$ are ordinary for every $\gamma \in \S{G}$. The assertion holds, because the union $\bigcup_\gamma \gamma\S{D}_{\vec{z}}$ is open and dense in $\S{X}$.
\qed

\subsubsection{The Weak Graph Representation Theorem}

Let $\omega: \S{G_H} \rightarrow \S{X}/\S{G}$ be the bijective isometry defined in Corollary \ref{Corollary:GRT}. A graph $Z \in \S{G_H}$ is ordinary, if there is an ordinary  representation $\vec{z} \in Z$. In this case, all representations of $Z$ are ordinary. The following result is an immediate consequence of Corollary \ref{cor:generic-ordinary}.

\begin{corollary}
A generic graph $Z \in \S{G_H}$ is ordinary.
\end{corollary}

The Weak Graph Representation Theorem describes the shape of a graph edit kernel space from a generic viewpoint.  
\begin{theorem}[Weak Graph Representation Theorem]\label{theorem:wgrt}
Suppose that $\args{\S{G_H}, \delta}$ is a graph edit kernel space. For each ordinary graph $Z \in \S{G_H}$ there is an injective map $\mu: \S{G_H} \rightarrow \S{X}$ into a Euclidean space $\args{\S{X}, \norm{\cdot}}$ such that 
\begin{enumerate}
\item $\delta(Z, X) = \norm{\mu(Z) - \mu(X)}$ for all $X \in \S{G_H}$.
\item $\delta(X, Y) \leq \norm{\mu(X) - \mu(Y)}$  for all $X \in \S{G_H}$.
\item The closure $\S{D}_\mu =  \cl\args{\mu(\S{G_H})}$ is a convex polyhedral cone in $\S{X}$.
\item We have $\S{D}_\mu^\circ \subsetneq \mu(\S{G_H}) \subsetneq \S{D}_\mu$.
\end{enumerate} 
\end{theorem}

\noindent
\proof
Let $\omega: \S{G_H} \rightarrow \S{X}/\S{G}$ be the bijective isometry defined in Corollary \ref{Corollary:GRT}.

\setcounter{part_counter}{0}
\begin{part}
Suppose that $Z \in \S{G_H}$ is a graph with $\vec{z} \in \omega(Z) = \bracket{\vec{z}}$. Since $Z$ is ordinary so is $\vec{z}$. Let 
$\S{D}_\mu = \S{D}_{\vec{z}}$ be the Dirichlet fundamental domain centered at $\vec{z}$, and $\S{F}_{\vec{z}} \subset \S{D}_\mu$ is a fundamental set. 
\end{part}

\begin{part}
The fundamental set $\S{F}_{\vec{z}}\subseteq \S{X}$ induces a bijection $f: \S{F}_{\vec{z}}\rightarrow \S{X}/\S{G}$ that maps each element $\vec{x}$ to its orbits $\bracket{\vec{x}}$. Then the map 
\[
\mu: \S{G_H} \rightarrow \S{X}, \quad X \mapsto f^{-1}(\omega(X))
\]
is injective as a composition of injective maps.
\end{part}

\begin{part}\label{part:wgrt:prop1}
We show the first property. Let $X$ be a graph. Then from Corollary \ref{Corollary:GRT} follows
\[
\delta(Z,X) = \min \cbrace{\norm{\vec{z}' -\vec{x}}\,:\, \vec{z}' \in \omega(Z)},
\]
where $\vec{x} = \mu(X)$. Since $\S{F}_{\vec{z}}$ is subset of the Dirichlet fundamental domain $\S{D}_{\vec{z}}$, we have
\[
\delta(Z,X) = \norm{\vec{z} -\vec{x}}.
\]
This shows the assertion.
\end{part}

\begin{part} We show the second property.  From the second part of this proof follows $\mu(X) = f^{-1}(\omega(X))$. This implies $\mu(X) \in \omega(X)$. Thus, $\mu$ maps every $X$ to exactly one  representation $\vec{x} \in \omega(X)$. Then the assertion follows from Corollary \ref{Corollary:GRT}.
\end{part}

\begin{part}
We have $\mu(\S{G_H}) = \S{F}_{\vec{z}}$. Then the third and fourth property follow from Prop.~\ref{proof:properties-of-DFD}.
\end{part}
\qed

\medskip

We call the map $\mu: \S{G_H} \rightarrow \S{X}$ an alignment of $\S{G_H}$ along $Z$. The polyhedral cone $\S{D}_\mu$ is the Dirichlet fundamental domain centered at $\mu(Z)$. Note that an alignment along $Z$ is not unique. 

\begin{figure}
\centering
\includegraphics[width=0.8\textwidth]{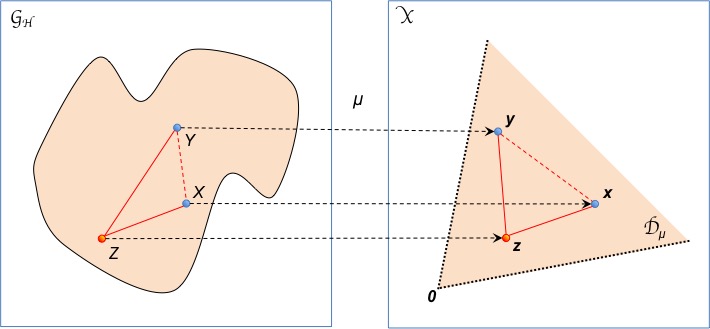}
\caption{Illustration of the Weak Graph Representation (WGR) Theorem. Suppose that $\mu:\S{G_H} \rightarrow \S{X}$ is an alignment along a generic graph $Z$. The box on the left shows the graph space $\S{G_H}$ together with graphs $X, Y, Z \in \S{G_H}$. The box on the right depicts the image $\mu(\S{G_H})$ as a region of the Euclidean space $\S{X}$ together with the images $\vec{x} = \mu(X)$, $\vec{y} = \mu(Y)$, and $\vec{z} = \mu(Z)$. Property (1) of the WGR Theorem states that $\mu$ is isometric with respect to $Z$. Distances are preserved for $\delta(Z, X) = \norm{\vec{z}-\vec{x}}$ and $\delta(Z, Y) = \norm{\vec{z}-\vec{y}}$ as indicated by the respective solid red lines.  Property (2) of the WGR Theorem states that $\delta(X, Y) \leq \norm{\vec{x} - \vec{y}}$ as indicated by the shorter dashed red line connecting $X$ and $Y$ in $\S{G_H}$ compared to the longer dashed red line connecting the images $\vec{x}$ and $\vec{y}$. Properties (3) and (4) of the WGR Theorem state that the closure $\S{D}_\mu$ of the image $\mu(\S{G_H})$ in $\S{X}$ (right box) is a convex polyhedral cone. Points of $\S{D}_\mu$ without pre-image in $\S{G_H}$ are boundary points as indicated by the holes in the dotted black line. Since a polyhedral cone can be regarded as a set of rays emanating from the origin, the side opposite of $\vec{0}$ of the Dirichlet fundamental domain $\S{D}_\mu$ is unbounded.}
\label{fig:alignment}
\end{figure}

The first property of the Weak Graph Representation Theorem states that there is an isometry with respect to a generic graph $Z$ into some Euclidean space. The second property  states that the alignment $\mu$ is an expansion of the graph space. Properties (3) and (4) say that the image of an alignment along a generic graph is a dense subset of a convex polyhedral cone. A polyhedral cone is the intersection of finitely many half-spaces. Figure \ref{fig:alignment} illustrates the statements of Theorem \ref{theorem:wgrt}. 

According to Theorem \ref{theorem:wgrt} an alignment $\mu$ along a generic graph $Z$ is an isometry with respect to $Z$, but generally an expansion of the graph space. 
Next, we are interested in convex subsets $\S{U} \subseteq \S{G_H}$ such that $\mu$ is isometric on $\S{U}$, because these subsets have the same geometrical properties as their convex images $\mu(\S{U})$ in the Euclidean space $\S{X}$ by isometry. To characterize such subsets, we introduce the notion of \emph{cone circumscribing a ball} for both metric spaces, the Euclidean space $\args{\S{X}, \norm{\cdot}}$ and the graph kernel edit space $\args{\S{G}, \delta}$.
\begin{definition}
Suppose that the metric space $(\S{M}, d)$ is either a Euclidean space or graph edit kernel space. Let $z \in \S{M}$ and let $\rho > 0$. A cone circumscribing a ball $\S{B}(z, \rho)$ is a subset of the form 
\[
\S{C}(z,\rho) = \cbrace{x \in \S{M} \,:\, \exists\, \lambda > 0 \text{ s.t. } \lambda x \in \S{B}(z, \rho)}.
\]
\end{definition}

The next results states that an alignment along an ordinary graph induces a bijective isometry between cones circumscribing sufficiently small balls.
\begin{theorem}\label{theorem:conic-isometry}
An alignment $\mu: \S{G_H} \rightarrow \S{X}$ along an ordinary graph $Z$ can be restricted to a bijective isometry from $\S{C}(Z,\rho)$ onto $\S{C}(\vec{z}, \rho)$ for all $\rho$ such that
 \[
 0 < \rho \leq \rho^* = \frac{1}{2} \min\cbrace{\norm{\vec{z} - \vec{x}} \,:\, \vec{x} \in \partial \,\S{D}_\mu},
 \]
 where $\vec{z} = \mu(Z)$.
\end{theorem}

\noindent
\proof
\setcounter{part_counter}{0}
\begin{part}
According to Corollary \ref{Corollary:GRT}, we have $\S{G_H} \cong \S{X}/\S{G}$. 
The group $\S{G}$ is a discontinuous group of isometries. Suppose that $\vec{z} \in Z$. Since $Z$ is ordinary, the isotropy group $\S{G}_{\vec{z}}$ is trivial. 
Then the natural projection $\pi:\S{X} \rightarrow \S{X}/\S{G}$ induces an isometry from $\S{B}(\vec{z}, \rho)$ onto $\S{B}(\pi(\vec{z}), \rho)$ for all $\rho$ such that 
\begin{align}\label{eq:proof:conic-isometry:rho}
0 < \rho \leq \frac{1}{4} \min \cbrace{\norm{\vec{z} - \gamma \vec{z}} \,:\, \gamma \in \S{G} \setminus \cbrace{\varepsilon}}
\end{align}
by \cite{Ratcliffe2006}, Theorem 13.1.1. This implies an isometry from $\S{B}(\vec{z}, \rho)$ onto $\S{B}(Z, \rho)$.
\end{part}

\begin{part}
Since $Z$ is ordinary, we have $\vec{z} \in \S{D}_\mu^\circ$ by Prop.~\ref{proof:properties-of-DFD}. The ball $\S{B}(\vec{z}, \rho)$ is contained in the open set $\S{D}_\mu^\circ$ for every radius $\rho$ satisfying eq.~\eqref{eq:proof:conic-isometry:rho}. To see this, we assume that $\S{B}(\vec{z}, \rho)$ contains a boundary point $\vec{x}$ of $\S{D}_\mu$. Then there is a $\gamma \in \S{G} \setminus \cbrace{\varepsilon}$ such that 
\[
\norm{\vec{x} - \gamma\vec{z}} = \norm {\vec{x} - \vec{z}} \leq \rho.
\]
We have 
\begin{align*}
\norm{\vec{z} -  \gamma\vec{z}} &= \norm{\vec{z} - \vec{x} + \vec{x} -  \gamma\vec{z}}\\
&\leq \norm{\vec{z} - \vec{x}} + \norm{\gamma\vec{z} - \vec{x}}\\
&\leq 2 \rho.
\end{align*}
Since $\rho$ satisfies eq.~\eqref{eq:proof:conic-isometry:rho}, we obtain a chain of inequalities of the form 
\[
\frac{1}{2} \norm{\vec{z} -  \gamma\vec{z}}  \leq \rho \leq \frac{1}{4}\norm{\vec{z} -  \gamma\vec{z}}.
\]
This chain of inequalities in invalid, because $\vec{z}$ is ordinary and $\gamma \neq \varepsilon$. From the contradiction follows $\S{B}(\vec{z}, \rho) \subseteq \S{D}_\mu^\circ$.
\end{part}

\begin{part} Let $\rho \in \;] 0, \rho^*]$. We show that $\S{C}(\vec{z}, \rho) \subset \S{D}_\mu^\circ$. The cone $\S{C}(\vec{z}, \rho)$ is contained in $\S{D}_\mu$ due to part one of this proof and convexity of $\S{D}_\mu$. Suppose that there is a point $\vec{x} \in \S{C}(\vec{z}, \rho) \cap \partial \S{D}_\mu$. Then there is a $\gamma \in \S{G} \setminus \cbrace{\varepsilon}$ such that $\vec{x}$ lies on the hyperplane $\S{H}$ separating the Dirichlet fundamental domains $\S{D}_{\mu}$ and $\gamma \S{D}_{\mu}$. 
Consider the ray $\S{L}_{\vec{x}} ^+= \cbrace{\lambda \vec{x} \,:\, 0 \leq \lambda} \subset \S{C}(\vec{z}, \rho)$. Two cases can occur: (1) either $\S{L}_{\vec{x}}^+ \subset \S{H}$ or (2)  $\S{L}_{\vec{x}}^+ \cap \S{H} = \cbrace{\vec{x}}$.  The first case contradicts that $\S{B}(\vec{z}, \rho)$ is in the interior of $\S{D}_\mu$, because the ray $\S{L}_{\vec{x}} ^+$ passes through $\S{B}(\vec{z}, \rho)$. The second case contradicts convexity of $\S{D}_\mu$. Thus we proved $\S{C}(\vec{z}, \rho) \subset \S{D}_\mu^\circ$.
\end{part}

\begin{part}
Let $\rho \in \;] 0, \rho^*]$ and $X, Y \in \S{C}(Z, \rho)$. Then there are positive scalars $a, b > 0$ such that $X' = aX$ and $Y' = bY$ are contained in $\S{B}(Z, \rho)$ by definition of $\S{C}(Z, \rho)$. Suppose that $\vec{x}' \in X'$ and $\vec{y'} \in Y'$ are representations of $X'$ and $Y'$ such that  $\vec{x}', \vec{y}' \in \S{B}(\vec{z}, \rho) \subseteq \S{D}_\mu^\circ$. Since $\S{G}$ is a subgroup of the general linear group, we have $\vec{x}' = a \vec{x}$ and $\vec{y}' = b \vec{y}$ with $\vec{x} \in X$ and $\vec{y} \in Y$. From part three of this proof follows that $\vec{x} = \vec{x}'/a$ and $\vec{y} = \vec{y}'/b$ are also contained in $\S{D}_\mu^\circ$. Applying Prop.~\ref{prop:positively-homogeneous} yields
\begin{align*}
\kappa(X', Y') &= (a\vec{x})^T(b\vec{y}) = ab \args{\vec{x}^T\vec{y}} = ab \cdot \kappa(X, Y)
\end{align*}
This implies $\delta(X,Y) = \norm{\vec{x} - \vec{y}}$. This shows a bijective isometry from $\S{C}(Z, \rho)$ onto $\mu(\S{C}(Z, \rho))$.
\end{part}

\begin{part} We show that $\mu(\S{C}(Z, \rho)) = \S{C}(\vec{z}, \rho)$. From part four of this proof follows that $\mu(\S{C}(Z, \rho)) \subseteq \S{C}(\vec{z}, \rho)$. It remains to show that 
 $\S{C}(\vec{z}, \rho) \subseteq \mu(\S{C}(Z, \rho))$. Let $\vec{x} \in  \S{C}(\vec{z}, \rho)$. Then there is a scalar $a>0$ such that $\vec{x}'= a\vec{x}$ is in the ball $\S{B}(\vec{z}, \rho) \subset \S{C}(\vec{z}, \rho)$. From the first part of the proof follows that there is a graph $X' \in \S{B}(X', \rho) \subset \S{C}(Z, \rho)$ with $\vec{x}' \in X'$. By definition of $\S{C}(Z, \rho)$, we have $X = X'/a$ is also in $\S{C}(Z, \rho)$. We need to show that $\mu(X) = \vec{x}$. From the third part of this proof follows that $\S{C}(\vec{z}, \rho) \subset \S{D}_\mu^\circ$.  This implies that $\vec{x} \in \S{D}_\mu^\circ$. From Prop.~\ref{proof:properties-of-DFD} follows that there is no other representation $\vec{x}' \in X$ contained in $\S{D}_\mu$. Since $\mu$ is surjective onto $\S{D}_\mu^\circ$ according to the Weak Graph Representation Theorem, we have $\mu(X) = \vec{x}$. This shows that $\S{C}(\vec{z}, \rho) \subseteq \mu(\S{C}(Z, \rho))$. 
\end{part}
\qed

\medskip

\begin{figure}
\centering
\includegraphics[width=0.6\textwidth]{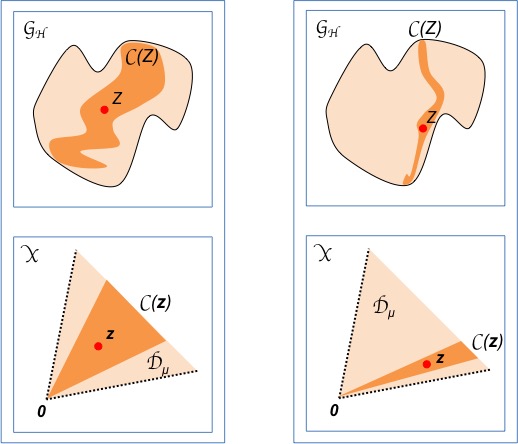}
\caption{Illustration of Theorem \ref{theorem:conic-isometry}. The two columns represent alignments $\mu$ along different graphs $Z$ into the Euclidean space $\S{X}$. The restriction of $\mu$ to the isometry cone $\S{C}(Y)$ is an isometric isomorphism into the (round) hypercone $\S{C}(\vec{z})$, where $\vec{z} = \mu(Z)$. is the image of graph $Z$. The isometric and isomorphic cones are shaded in dark orange. The hypercone $\S{C}(\vec{z})$ is wider the more $\vec{z}$ is centered within its Dirichlet fundamental domain $\S{D}_\mu$.}
\label{fig:isometry}
\end{figure}

The maximum radius $\rho^*$ in Theorem \ref{theorem:conic-isometry} is half the minimum distance of $\vec{z}$ from the boundary of its Dirichlet fundamental domain $\S{D}_\mu$. The circular cone $\S{C}(\vec{z}, \rho^*)$ in $\S{X}$ is wider the more centered $\vec{z}$ is within its Dirichlet fundamental domain $\S{D}_\mu$. Then by isometry, the cone $\S{C}(Z, \rho^*)$ in $\S{G_H}$ is also wider. Note that for every generic graph $Z$, the circular cone $\S{C}(\vec{z}, \rho^*)$ never collapses to a single ray. Figure \ref{fig:isometry} visualizes Theorem \ref{theorem:conic-isometry}.

A direct implication of the previous discussion is a correspondence of basic geometrical concepts between graph edit kernel spaces and their images in Euclidean spaces via alignment along an ordinary graph. The next result summarizes some correspondences.
\begin{corollary}
Let $\mu:\S{G_H} \rightarrow \S{X}$ be an alignment along an ordinary graph $Z \in \S{G_H}$. Then the following statements hold for all $X \in \S{G_H}$:
\begin{enumerate}
\item $\kappa(Z, X) = \mu(Z)^T \mu(X)$ for all $X \in \S{G_H}$.
\item $\ell(X) = \norm{\mu(X)}$
\item $\sphericalangle(Z, X) = \sphericalangle\args{\mu(Z), \mu(X)}$.
\item $Z$ and $X$ are orthogonal $\; \Leftrightarrow \;$ $\mu(Z)$ and $\mu(X)$ are orthogonal. 
\item $Z$ is orthogonal to a subset $\S{U} \subseteq \S{G_H}$ $\; \Rightarrow \;$ $\mu(Z)$ is orthogonal to $\mu(\S{U})$.
\end{enumerate}
\end{corollary}

\section{Discussion}
This contribution studies the geometry of graph edit kernel spaces. Results presented in this paper serve as a basis for statistical data analysis on graphs. The main result is the Graph Representation Theorem. It states that under mild assumptions graphs are points of a geometrical space, called orbit space. Orbit spaces are well investigated and easier to explore than the original graph edit kernel space.  Consequently, we derived a number of results from orbit spaces useful for statistical data analysis on graphs and translated them to graph edit kernel spaces. 

In the remainder of this section, we conclude with indicating the significance and usefulness of the results for statistical pattern recognition on graphs. 

Graph edit kernels and graph metrics induced by edit kernels are used in numerous applications. Consequently, there is ongoing research on devising graph matching algorithms for computing graph edit kernels and their induced metric \cite{Almohamad1993,Caetano2007,Cho2010,Cour2006,Gold1996,Leordeanu2005,Leordeanu2009,Schellewald2005,Umeyama1988,Wyk2002,M.Zaslavskiy2009,Zhou2012}.

The notion of angle, length, and orthogonality together with the Graph Representation Theorem and its weak version are useful for a geometric interpretation of linear classifiers generalized to graph edit kernel spaces \cite{Jain2014a,Jain2014b}. 

One of the most fundamental statistic is the concept of mean of a random sample of graphs. The Weak Graph representation Theorem, Theorem \ref{theorem:geodesic-space} and Theorem \ref{theorem:conic-isometry} partly in conjunction with results from \cite{Bhattacharya2012} are useful for addressing the following issues:
\begin{enumerate}
\item Existence of a sample mean of graphs.
\item Uniqueness of a sample mean of graphs. 
\item Strong consistency of sample mean of graphs.
\item Midpoint property of a mean of two graphs.
\item Vectorial characterization of sample mean of graphs.
\end{enumerate}

\end{document}

%% file: config.tex
\usepackage{makeidx}
\usepackage{graphicx}
\usepackage{amssymb}
\usepackage{amsmath}
\usepackage{amscd}
\usepackage{tikz}
\usetikzlibrary{matrix,arrows}
\usepackage{multicol}

\newtheorem{definition}{Definition}[section]
\newtheorem{theorem}[definition]{Theorem}
\newtheorem{corollary}[definition]{Corollary}
\newtheorem{proposition}[definition]{Proposition}

\newcommand{\qed}{\hspace*{\fill} $\blacksquare$}
\newcommand{\proof}{Proof:\ }

\newcommand{\commentout}[1]{}

\newcommand{\R}{\mathbb{R}}                    

\newcommand{\abs}[1]{\mathop{\left\lvert #1 \right\rvert}} 
\newcommand{\args}[1]{\mathop{\left( #1 \right)}} 

\newcommand{\norm}[1]{\mathop{\left\lVert #1 \right\rVert}}
\newcommand{\cbrace}[1]{\mathop{\left\{ #1 \right\}}}
\newcommand{\bracket}[1]{\mathop{\left[ #1 \right]}}

\newcommand{\argsS}[2]{\mathop{\left( #1 \right)#2}} 

\newcommand{\normS}[2]{\mathop{\left\lVert #1 \right\rVert#2}}

\DeclareMathOperator{\cl}{cl}                  

\DeclareMathOperator{\id}{id}                  		

\renewcommand{\S}[1]{{\mathcal{#1}}}           	
\def\vec#1{\mathchoice{\mbox{\boldmath$\displaystyle#1$}}
{\mbox{\boldmath$\textstyle#1$}}
{\mbox{\boldmath$\scriptstyle#1$}}
{\mbox{\boldmath$\scriptscriptstyle#1$}}}

\renewenvironment{cases}{%
\left\{\begin{array}{c@{\quad : \quad}l}}%
{%
\end{array}\right.}

\newcounter{part_counter}
\renewenvironment{part}{
\refstepcounter{part_counter}

\medskip

\noindent
\textbf{\arabic{part_counter}.} 
}%

\newcounter{algorithm_counter}
{
\\[-1.5ex]
\rule{\textwidth}{\arrayrulewidth}
\end{footnotesize}
\end{minipage}
\setlength{\parindent}{\parindent}
}

{
\\[-1.5ex]
\rule{\textwidth}{\arrayrulewidth}
\end{footnotesize}
\end{minipage}
\setlength{\parindent}{\parindent}
}